\documentclass[letterpaper, 10 pt, conference]{ieeeconf}  

\IEEEoverridecommandlockouts                              
\usepackage[final]{microtype}
\usepackage[pdftex]{graphicx}
\usepackage[cmex10]{amsmath}
\usepackage{amssymb}
\usepackage[ruled,vlined]{algorithm2e}
\usepackage{array}
    
\usepackage{multirow}
\usepackage{arydshln}
\usepackage[caption=false,font=footnotesize]{subfig}
\usepackage{url}
\usepackage{booktabs}
\usepackage{tabularx}
\usepackage{cite}
\usepackage{dashrule}
\usepackage{xspace} 
\usepackage{fixltx2e}
\usepackage{graphicx}
\usepackage{import}

\usepackage{xcolor}
    \definecolor{dark-red}{rgb}{0.4,0.1,0.1}
    \definecolor{dark-blue}{rgb}{0.15,0.15,0.5}
    \definecolor{medium-blue}{rgb}{0.0,0.0,0.8}
    \definecolor{mycolor1}{rgb}{1,0.2,0.2}
\usepackage{hyperref}
\usepackage{dcolumn}
    \newcolumntype{W}[1]{D{,}{,}{#1}}
    \newcolumntype{M}[1]{D{}{}{#1}}

\newcolumntype{+}{>{\global\let\currentrowstyle\relax}}
\newcolumntype{^}{>{\currentrowstyle}}

\hypersetup{
    colorlinks,
    citecolor=black,
    filecolor=black,
    linkcolor=black,
    urlcolor=black
}

\usepackage{flushend}
\usepackage{fancyhdr}
\definecolor{colG}{rgb}{0,0.8,0}
\definecolor{colB}{rgb}{0,0.0,0.7}
\definecolor{colR}{rgb}{1.0,0,0}

\newcommand{\eref}[1]{(\ref{#1})}

\newcommand{\fref}[1]{Fig.{\;}\ref{#1}}
\newcommand{\sref}[1]{Sec.{\;}\ref{#1}}

\title{\LARGE \bf
Unsupervised Spike Sorting Based on Discriminative Subspace Learning
}
\author{Mohammad Reza Keshtkaran~and Zhi Yang
\thanks{This work is supported by Minister of Education Tier-1 funding R-263-000-A47-112, Singapore A*STAR Public Sector funding R-263-000-699-305, and Young Investigator Award R-263-000-A29-133.}
\thanks{Mohammad Reza Keshtkaran and Zhi Yang are with the Department of Electrical and Computer Engineering, National University of Singapore, 117576, Singapore~
        (e-mail: \small\href[]{mailto:keshtkaran@u.nus.edu}{keshtkaran@u.nus.edu})%
}}

\newcommand{\apriori}{\emph{a priori}\xspace}
\newcommand{\kmeans}{\mbox{$k$-means}\xspace}
\newcommand{\invivo}{\emph{in-vivo}\xspace}
\newcommand{\Sw}{S_\mathrm{w}}
\newcommand{\Sb}{S_\mathrm{b}}
\newcommand{\Ck}{\mathcal{C}_\mathrm{k}}

\begin{document}

\maketitle
\thispagestyle{fancy}

\begin{abstract}
Spike sorting is a fundamental preprocessing step for many neuroscience studies which rely on the analysis of spike trains. In this paper, we present two unsupervised spike sorting algorithms based on discriminative subspace learning. The first algorithm simultaneously learns the discriminative feature subspace and performs clustering. It uses histogram of features in the most discriminative projection to detect the number of neurons. The second algorithm performs hierarchical divisive clustering that learns a discriminative 1-dimensional subspace for clustering in each level of the hierarchy until achieving almost unimodal distribution in the subspace.
The algorithms are tested on synthetic and \invivo data, and are compared against two widely used spike sorting methods. The comparative results demonstrate that our spike sorting methods can achieve substantially higher accuracy in lower dimensional feature space, and they are highly robust to noise. Moreover, they provide significantly better cluster separability in the learned subspace than in the subspace obtained by principal component analysis or wavelet transform.
\end{abstract}

\section{Introduction}
Extracellular recording is a method of measuring neuronal activity, which is commonly used by neuroscientists to study brain functions. Neural signals are recorded by inserting microelectrodes into the brain tissue which picks up local field potentials, the action potentials (also called spikes) from a few surrounding neurons, and noise. For obtaining \mbox{(multi-)unit} activity, the signal is often filtered in \mbox{300--5000~Hz} frequency band and spikes are identified through using a spike detection method. In many neuroscience studies, it is necessary to sort the spikes after detection so that the spikes generated by an individual unit fall into one cluster. The spike sorting stage is fundamental to the neuroscience studies which involve the analysis of spike rates, spike time synchrony, and \mbox{inter-spike interval \cite{lewicki_review_1998,gibson_spike_2012}}.

Common spike sorting methods involve detecting neural spikes, extracting and selecting features from detected spike waveforms, detecting the number of neurons, and assigning the spikes to their originating neurons \cite{lewicki_review_1998}. For spike detection, methods based on absolute value thresholding, nonlinear energy operator, and wavelet have been widely used \cite{gibson_spike_2012}.

Among different feature extraction methods, principal component analysis (PCA) and discrete wavelet transforms (DWT) are most commonly used \cite{lewicki_review_1998,adamos_performance_2008,gibson_spike_2012}. PCA projects the spikes to a set of orthogonal basis vectors that represent the largest variance of the data. In wavelet-based methods the spikes are decomposed into wavelets and the decomposition coefficients are used as features. A good feature extraction method should retain the most useful information for discriminating different spike shapes in a reasonably low dimension \cite{gibson_spike_2012}. Some efforts have been done to come up with better feature extraction for spike sorting including methods based on waveform derivatives \cite{yang_improving_2009}, Laplacian eigenmaps \cite{chah_automated_2011}, wavelet optimization \cite{shalchyan_spike_2012}, and Fourier transform \cite{yang_robust_2013}. Although these methods can provide more discriminative features with reasonable dimension reduction, they do not necessarily seek for the most discriminative subspace for clustering \cite{ding_adaptive_2007}. Hence, the clusters which appear inherently separable in some discriminative subspace may overlap if projected using conventional features extraction methods. Such cluster overlaps increases the misclassification, may lead to incorrect detection of number of clusters, hindering reliable clustering. Therefore, feature learning methods which seek for the most discriminative subspace is expected to provide an optimal cluster separation thus improving the clustering performance in spike sorting.

Depending on the proximity of an electrode to the surrounding neurons, the recording may contain several spike waveforms generated by different neurons. In practice, the number of the contributing neurons (i.e. clusters) is not known \apriori and needs to be detected from the data. Incorrect selection of the number of the clusters heavily affects the performance of the clustering algorithm. One way to set the number of neurons is through visual inspection of spike waveforms by an expert observer. This method makes the spike sorting non-automatic and is prone to human error. It is generally desired that the algorithm detects the number of the clusters in an unsupervised manner so as to eliminate the user intervention \cite{lewicki_review_1998,gibson_spike_2012}.

Automatic detection of the number of neurons often depends on the method used for clustering. Some common clustering methods used for spike sorting include \kmeans,  mixture models, neural networks, superparamagnetic clustering (SPC), and self-organizing maps \cite{lewicki_review_1998, kim_neural_2000, quiroga_unsupervised_2004, yang_robust_2013}. Several approaches based on Bayesian/Akaike information criterion, gap statistics, and statistical test for Gaussian distribution have been proposed to learn the number of clusters for \kmeans and mixture model \cite{novak_identifying_2009, jain_data_2010}. In SPC the number of clusters are detected by sweeping the temperature and choosing the clusters which are bigger than a certain size \cite{quiroga_unsupervised_2004}. In general, various combinations of feature extraction and clustering have been applied to spike sorting with different \mbox{levels of success \cite{adamos_performance_2008,gibson_spike_2012}.}

In this paper, we propose two unsupervised spike sorting algorithms based on discriminative subspace learning to extract low dimensional and most discriminative features from the spike waveforms, and perform clustering. The core part of the algorithms involves iterative subspace selection using linear discriminant analysis (LDA) and clustering using \kmeans. The first algorithm directly uses the core part to simultaneously learn the most discriminative subspace and perform clustering. It uses histogram of features in the most discriminative projection to detect the number of neurons.  The second algorithm performs
hierarchical divisive clustering that learns a discriminative 1-dimensional subspace for clustering in each level of the hierarchy until achieving almost unimodal distribution in the subspace. The algorithms are tested on synthetic and \invivo neural data, and their performances are compared with two commonly used spike sorting methods PCA-$k$means and \texttt{Wave\_clus} \cite{quiroga_unsupervised_2004} with regard to sorting accuracy, robustness to noise, and separation of the clusters.

 The rest of this paper is organized as follows. \sref{sec:method} describes the two proposed spike sorting algorithms based on discriminative subspace learning. \sref{sec:results} gives the comparative results on synthetic and \invivo data. And \sref{sec:conclusion} concludes the paper.

\section{Method} \label{sec:method}
In this work, we assume that the spikes are already detected through one of the common approaches, aligned to their peak, and stored in $X$.
The following section describes the discriminative subspace learning method which is the core building block of the two proposed algorithms. The proposed algorithms for detecting the number of neurons and sorting the spikes are presented in \sref{sec:algo1} and \sref{sec:algo2}.

\subsection{Discriminative Subspace Learning using LDA and \mbox{\kmeans}} \label{sec:LDAKm}
Most of the feature extraction and dimensionality reduction techniques that have been used for spike sorting give a projection subspace which is not necessarily the most discriminative one. Feature extraction (followed by dimension reduction) is a crucial stage which determines the quality of the next stage clustering. Thus, feature extraction should transform the data in such a way that similar data points are close to each other while dissimilar ones are well-separated from each other. For this purpose, we utilize a discriminative subspace learning technique using iterative application of LDA and \kmeans.

Suppose the spikes are stored in $X_{(m\times n)}$ where $m$ is the number of samples stored for each spike waveform and $n$ is the total number of the detected spikes.
We are interested in finding a discriminative projection matrix $W_{(m\times d)}$ to transform the spike waveforms to a $d$-dimensional ($d<m$) feature space $Y_{(d\times n)}$ so that the potential clusters have maximum separability:
\begin{IEEEeqnarray}{C}
    Y = W^T X . \label{eq:proj}
\end{IEEEeqnarray}
Clustering is then performed in the subspace $Y$. For now, it is assumed that the number of clusters is known and is indicated by $K$; we will discuss the selection of $K$ in \sref{sec:algo1} and \sref{sec:algo2}. Let $L_{(n\times K)}$ be the cluster indicator matrix which assigns each data point (i.e. spike) $x_i$ to its corresponding cluster $\Ck$ so that $L_{i,k}=1$ if $x_i \in \Ck$, and $L_{i,k}=0$ otherwise.
The cluster density and separability can be quantized by within- and between-class scatter matrices respectively defined as
\begin{IEEEeqnarray}{lrCl}
    &\Sw &=& \sum_{k=1}^{K}{\sum_{x_i\in \Ck}{(x_i-\mu_k)(x_i-\mu_k)^T}} \nonumber \\
        &&=& (X - M L^T)(X - M L^T)^T , \\
\noalign{\noindent and}\nonumber \\
    &\Sb &=& \sum_{k=1}^{K}{n_k\mu_k\mu_k^T} = M L^T L M^T ,
\end{IEEEeqnarray}
where $\mu_k$ is the center of $\Ck$, $M = (\mu_1\cdots\mu_K)$, and $n_k$ is the number of points in $\Ck$.
To achieve a high cluster separability, the within-class scatter should be small and/or the between-class scatter should be large. This leads to the following optimization problem:
\begin{IEEEeqnarray}{C}
    \max_{W,L}{\mathrm{Tr}\frac{W^T \Sb W}{W^T \Sw W}} \label{eq:LDAkm}
\end{IEEEeqnarray}
Some solutions to this problem have been presented in \cite{ding_adaptive_2007}. With $W$ fixed, \eref{eq:LDAkm} becomes a \kmeans clustering in feature space $Y$; with $L$ fixed, it can be solved for $W$ by LDA. A straightforward method is to fix $W$ (initialized by PCA) and use \kmeans to obtain $L$, and using the updated $L$ then perform LDA to update $W$ and iterate this procedure until convergence (i.e. $L$ remains the same between two iterations). This method is referred to as \emph{LDA-Km} through which, the data are simultaneously clustered while the discriminative subspaces are selected.
Since \kmeans is used as a building block of the algorithm, the limitations inherent in \kmeans such as local convergence and sensitivity to initialization also exist here. However, various methods exist to mitigate these problems such as repeating \kmeans clustering with different initializations and picking the one which leads to the least intra-cluster distance \cite{jain_data_2010}.

\subsection{Proposed Algorithm~1} \label{sec:algo1}
The first algorithm is based on the direct application of LDA-Km to detect the number of clusters and sort the spikes.
It is desired to discover in the data as many potential clusters as possible while avoiding over-clustering. In this work we propose a technique which uses the histogram of features in 1-dimensional discriminative subspace, and seeks the peaks of the histogram to infer the number of the clusters in the data. It is assumed that the clusters are denser at their centers and more spread at borders. This transforms to identifiable peaks in the histogram of the features in the most discriminative subspace. Therefore, the number of the peaks would be a reasonable indicator of the number of the clusters.

Simultaneous subspace learning and clustering with different values of $K$ leads to different features spaces. We look for a clustering with the biggest $K$ in which the number of the peaks in the histogram is equivalent to (or not less than) $K$. As so, we keep increasing $K$ and do the subspace learning until the number of the peaks appearing in the histogram becomes less than $K$ and remains unchanged (no well-separated cluster for bigger $K$). This method is summarized in Algorithm~\ref{algo:algo1}.

It should be noted that, the 1-dimensional discriminative subspace for calculating the histogram is obtained by performing LDA with the iterative trace ratio (ITR) solution \cite{wang_trace_2007}, which leads to better separation of the clusters in the subspace. Furthermore, similar to other density estimation methods, the parameters used for calculating the histogram such as bin width and smoothing factor, affect the estimation and may alter the number of the observed peaks. Several approaches can be utilized for adjusting these parameters for more reliable density estimation \cite{silverman_density_1986}.



\subsection{Proposed Algorithm~2} \label{sec:algo2}
Here, we propose a clustering algorithm based on divisive clustering scheme and use it for spike sorting. In this method, the data are initially partitioned into two clusters (using LDA-Km) in the most discriminative 1-dimensional subspace. The resultant clusters are marked as \emph{child}. After that, the distribution of samples in the obtained subspace is tested for unimodality using Anderson-Darling (AD) test (or similar tests). If the test score is higher than a predefined threshold, which indicates that the distribution is multimodal, each of the \emph{child} clusters are again partitioned into two clusters and marked as \emph{child}, otherwise they are merged into one cluster and marked as \emph{final}. The same process is then repeated for each \emph{child} cluster until all the clusters are marked as \emph{final}. In this process, clusters with sizes smaller than a predefined value, or sparse samples that lie far from cluster centers, can be neglected and assigned as outliers.

The benefit of this method compared with Algorithm~1 is that 1-dimensional subspace is used for clustering which leads to a faster convergence and better handling of clusters with different densities. Moreover, this algorithm has the ability to handle outliers. Since this is a hierarchical clustering approach, a drawback is that misclassified samples in the top levels will affect the clustering results in subsequent levels. The summary of the proposed algorithm is provided in Algorithm~\ref{algo:algo2}.

It should be noted that, the threshold controls how similar the distribution of the clusters should be to gaussian to be considered as \emph{final}. In general, a larger threshold value causes the algorithm to extract more spread clusters, whereas a smaller value results in the extraction of more compact and smaller clusters which may lead to over-clustering. Based on extensive tests on synthetic and real data, threshold values of 30--50 (AD test score) seem to provide a reasonable trade-off.

\section{Results} \label{sec:results}
Evaluation of spike sorting algorithms is often challenging due to the lack of ground truth \cite{lewicki_review_1998,gibson_spike_2012}. A common approach, however, is to use synthetic datasets, which would provide the ground truth as well as the flexibility to evaluate different characteristics of the spike sorting algorithm. Synthetic neural data is usually prepared by replicating spikes from a few spike waveform templates and adding them to a background noise mimicking neural noise, with the spike arrival time distributions similar to those of real spike trains.  In this work, we evaluate our algorithms using synthetic data and compare the results to two other spike sorting methods. Furthermore, we present qualitative clustering results using real \invivo data, in which case the performance of the algorithms can be qualitatively assessed by looking into cluster separation and inter-spike interval (ISI) histogram.

\begin{algorithm}[!t]
\label{algo:algo1}
\small
\caption{Spike Sorting using Discriminative Subspace Learning}

\KwIn{$X_{m,n}$: $n$ spikes, $m$ samples for each spike}
Initialize: $K \gets 1$, $P_1\gets 0$\;
\Repeat{$P_K=P_{K-1}<K$}{
        $K \gets K +1$\;
        do LDA-Km on $X$ to extract $K$ clusters $L^K$, and \\ features $Y^K$\;
        perform ITR \cite{wang_trace_2007} using $L^K$ to obtain 1-dimensional discriminative subspace of $X$\;
        estimate the histogram in the 1-dimensional subspace\;
        $P_K \gets$ number of peaks in the histogram\;
}
Cluster $X$ using $L^{K-1}$\;
\end{algorithm}

\begin{algorithm}[!t]
\label{algo:algo2}
\small
\caption{Spike Sorting using Discriminative Divisive Clustering}
\KwIn{$X_{m,n}$: $n$ spikes, $m$ samples for each spike}
\textbf{Define:} $\mathcal{S} = \{S_1, S_2, \cdots, S_I\}$\;
\textbf{Initialize:} $K \gets 2$, $\mathcal{S} \gets \{X\}$\;
\Repeat{All the spikes are assigned to clusters}{
\ForEach{$S_i \in \mathcal{S}$}
{do LDA-Km on $S_i$ to extract $K$ \emph{child} clusters ($S_i^j$)\;
    \eIf{ AD test score on $S_i$ $<$ Threshold}{
    Create a \emph{final} cluster from $S_i$
    }{
    \ForEach{$S_i^j$}{
    {
        \eIf{Size($S_i^j$) $>$ Minimum Cluster Size}
            {Add $\{S_i^j\}$ to $\mathcal{S}$}
        {Assign $S_i^j$ to the Outliers Cluster}
    }
    }
Remove ${S_i}$ from $\mathcal{S}$\;
}
}
}
\end{algorithm}

We compared the two proposed algorithm with \mbox{PCA-$k$means} and \texttt{Wave\_clus} based on criteria such as accuracy of spike sorting, dimensionality of feature space for effective clustering, and sensitivity to noise.

In PCA-$k$means, PCA is used for feature extraction and dimensionality reduction, and \kmeans is used for clustering. This method is supervised meaning that the number of clusters ($k$ in \kmeans) must be known, and parametric assuming the clusters are spherical in the PCA space.  \texttt{Wave\_clus} is an spike sorting method which uses DWT for feature extraction and SPC for clustering. It has been shown very practical for spike sorting since it can automatically choose the number of clusters (unsupervised), and does not impose any assumption on the shape of the clusters (nonparametric). In this method, Haar wavelet coefficients are used as features where the 10 coefficients with largest deviation from normality were selected as the input to the SPC algorithm. In many situations, SPC clusters a subset of the spikes and the remaining spikes are forced to be classified using a KNN classifier.
\begin{figure}[t]
\centering
\subfloat[][]{\includegraphics[width=0.19\textwidth,height=0.18\textwidth]{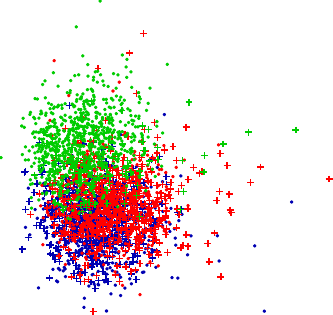}} \hfil
\subfloat[][]{\includegraphics[width=0.19\textwidth,height=0.18\textwidth]{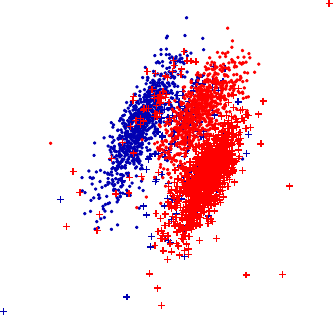}}
\\
\subfloat[][]{\includegraphics[width=0.19\textwidth,height=0.18\textwidth]{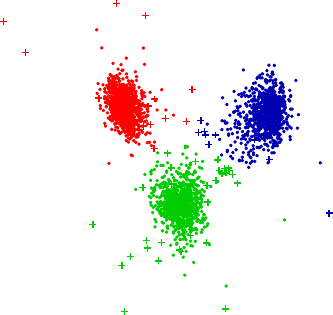}} \hfil
\subfloat[][]{\includegraphics[width=0.15\textwidth,height=0.18\textwidth]{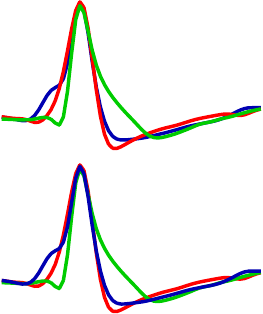}}
\caption{\label{fig:synth} Results on synthetic data (\texttt{C\_Difficult2\_noise02}). (a) PCA-$k$means where feature dimension is 10 (projection on the first two principal components is shown), and $K$ is manually set to 3. Accuracy = 60.6\%. (b) \texttt{Wave\_clus} where feature dimension is 10 (best 2-D DWT projection is shown); SPC fails to identify the third cluster for any temperature value. Accuracy = 58.9\%. (c) Proposed Algorithm~1 where feature dimension is 2 (most discriminative projections). The 3 clusters are automatically identified. Accuracy = 98.2\%. Misclassified spikes are shown as `+'. (d) Upper plot: Average spike waveforms of the clustering in (c). Lower plot: Average spike waveform of the sorted spikes using Algorithm~2. Accuracy = 98.3\%.}
\end{figure}

\begin{figure}[t]
\centering
\subfloat[][]{\includegraphics[width=0.3\columnwidth,height=0.3\columnwidth]{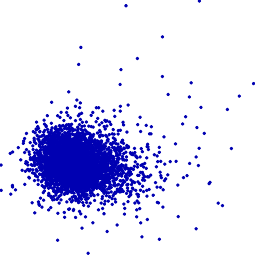}} \hfil
\subfloat[][]{\includegraphics[width=0.3\columnwidth,height=0.3\columnwidth]{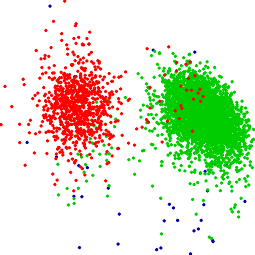}} \hfil
\subfloat[][]{\includegraphics[width=0.3\columnwidth,height=0.3\columnwidth]{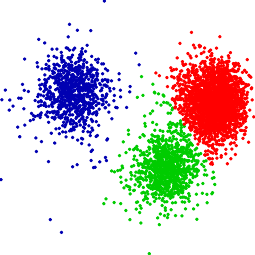}}  \\
\subfloat[][]{\includegraphics[width=0.25\columnwidth,height=0.2\columnwidth]{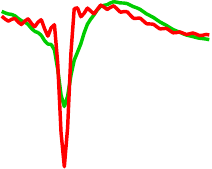}} \hfil
\subfloat[][]{\includegraphics[width=0.25\columnwidth,height=0.2\columnwidth]{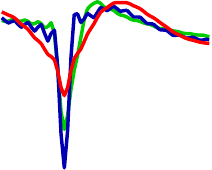}}  \qquad
\subfloat[][]{\includegraphics[width=0.25\columnwidth,height=0.2\columnwidth]{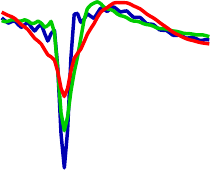}} \\
\subfloat[][]{\includegraphics[width=0.29\columnwidth,height=0.22\columnwidth]{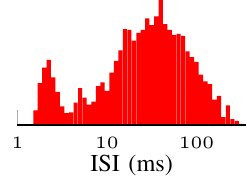} \hfil
\includegraphics[width=0.29\columnwidth,height=0.22\columnwidth]{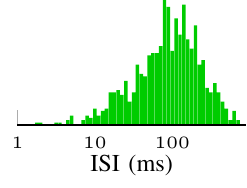} \hfil
\includegraphics[width=0.29\columnwidth,height=0.22\columnwidth]{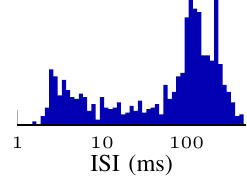}}

\caption{\label{fig:real} Comparative results on real data recorded from the rat hippocampus (\texttt{d1122101:1}). (a) Projection on the first two principal components. The clusters cannot be identified through PCA. (b) Best 2-D projection of DWT features used in \texttt{Wave\_clus} where feature dimension is 10. Only two clusters are identified (points in blue are assigned as outliers).  (c) Discriminative features learned by Algorithm~1, the three clusters are automatically identified. (d) and (e) show the average spike waveforms of the clustering in (b) and (c), respectively. (f) Average spike waveform of the sorted spiked using Algorithm~2. The three clusters are automatically identified and the results are highly similar to those of Algorithm~1. The histogram of the log inter-spike interval of spikes sorted by Algorithm~2, indicating the validity of the extracted clusters. Colors correspond to the spikes in (f).}
\end{figure}

For simulation with synthetic data, we have used eight challenging datasets provided by \cite{quiroga_unsupervised_2004}, which have been widely used in the literature to benchmark spike sorting algorithms. The datasets \texttt{C\_difficult1*} and \texttt{C\_difficult2*} are referred as Set~1 and Set~2, respectively. Each dataset contains spikes from three different neurons.
Simulations are carried out for different noise levels and feature dimension of 2--10 for PCA-$k$means and \texttt{Wave\_clus}; feature dimension was 2 for Algorithm~1, and 1 for Algorithm~2. The number of clusters is manually set to 3 for PCA-$k$means, while \texttt{Wave\_clus} and the proposed algorithms detect it automatically. Throughout our simulations, we used the default parameters of \texttt{Wave\_clus} and chose the best results on multiple runs of the algorithm. \texttt{Wave\_clus} correctly detected the number of clusters for all cases except for Set\,2\textbackslash$\sigma_\mathrm{n}{=}0.2$ where it only detected 2 clusters. The proposed algorithms detected the true number of clusters in all the cases. The results are shown in Table~\ref{tab:results}, where the first column is the dimension of the feature space, and the first row is the standard deviation of the noise ($\sigma_n$). Overlapping spikes are excluded in calculating the accuracies. It can be seen that the accuracy of PCA-$k$means and \texttt{Wave\_clus} degrades as the dimension of the features reduces. \texttt{Wave\_clus} (with forced clustering) performs better than PCA-$k$means especially when the noise level is high; however, the performance of both method degrade significantly as the noise level increases. Interestingly, the proposed methods give significantly higher sorting accuracies compared with the other methods and they are highly robust to noise.

\fref{fig:synth} shows the feature extraction and clustering results on the most challenging dataset  Set\;2\textbackslash${\sigma_\mathrm{n}{=}0.2}$ (\texttt{C\_Difficult2\_noise02}). As can be seen in \fref{fig:synth}(a)--(c), the 2-D subspace learned by Algorithm~1 provides significantly better separability of clusters compared with PCA and DWT. \fref{fig:synth}(d) shows the average waveforms of spikes sorted by Algorithm~1 (upper) and Algorithm~2 (lower).

\begin{table}[t]
\scriptsize
\smallskip\noindent
\newlength{\tcolsep}
\setlength{\tcolsep}{0.7em}
\caption{\label{tab:results} Comparative Results on Synthetic Data}
\resizebox{\columnwidth}{!}{
\begin{tabular}{@{}c@{}c@{}c@{}c@{}c@{}c@{}c@{}c@{}c@{}c@{}c@{}c@{}c@{}c@{}c@{}c@{}c@{}c@{}c@{}c@{}c@{}c@{}c@{}c@{}c@{}c@{}c@{}}
\toprule
Dim. & \hspace{\tcolsep} & Set && \multicolumn{7}{@{}c@{}}{PCA-$k$means (\%)} && \multicolumn{7}{@{}c@{}}{\texttt{Wave\_clus} (\%)} &&  \multicolumn{7}{@{}c@{}}{Algorithm~1 (\%)} \\
\cmidrule{1-1} \cmidrule{3-3} \cmidrule{5-11} \cmidrule{13-19} \cmidrule{21-27}

\multicolumn{3}{@{}c@{}}{$\sigma_{n}$:}  & \hspace{\tcolsep} & 0.05 & \hspace{\tcolsep} & 0.1 & \hspace{\tcolsep} & 0.15 & \hspace{\tcolsep} & 0.2 & \hspace{2\tcolsep} & 0.05 & \hspace{\tcolsep} & 0.1 & \hspace{\tcolsep} & 0.15 & \hspace{\tcolsep} & 0.2 & \hspace{2\tcolsep} & 0.05 & \hspace{\tcolsep} & 0.1 & \hspace{\tcolsep} & 0.15 & \hspace{\tcolsep} & 0.2 \\
\cmidrule{1-27}

\multirow{2}{*}{$2$}
       && 1 && 93.2 && 75.2 &&  50.8 && 40.9 &&  97.7 &&  89.0 && 63.1 && 62.2 && 99.6 && 99.4 &&  99.1 && 99.2 \\
       && 2 && 81.1 && 74.0 && 65.8 && 56.7 && 66.0 && 52.6 && 50.7 && 49.6 && 98.7 && 98.9 && 98.7 && 98.2 \\
\cdashline{1-19}[1pt/1pt]
\multirow{2}{*}{$4$}
       && 1 && 97.9 &&  84.7 && 64.7 && 53.5 && 98.7 && 96.9 && 93.0 && 86.5 && \\
       && 2 && 91.8 &&  80.8 && 72.7 && 63.3 && 93.3 &&  98.0 && 91.0 && 58.0 && \multicolumn{7}{c}{Algorithm~2$^\dagger$ (\%)} \\
\cdashline{1-19}[1pt/1pt] \cline{21-27}
\multirow{2}{*}{$10$}
       && 1 && 98.9 && 95.1 && 85.5 && 76.8 && 98.4 &&  98.95 && 95.4 && 86.5  &&  98.1 && 99.2 && 99.1 && 98.7 \\
       && 2 && 97.8 &&  91.3 && 83.0 && 60.6 && 97.3 &&  98.4 && 91.7 && 58.9 && 98.6 && 98.7 && 98.8 && 98.3 \\
\bottomrule
\end{tabular}
}
\vspace{0.5pt}

${\;}^{\dagger}$Feature dimension is 1.
\end{table}


The results of simulation using \invivo data is shown in \fref{fig:real}. The data is recorded from the rat hippocampus and is publicly available \cite{datasets}. We chose the recording \texttt{d1122101:1} which we found challenging for spike sorting. The spikes were extracted by negative amplitude thresholding at 3~RMS, and aligned to their peak values. In total, \mbox{4541~spikes} were detected and 64 samples were stored for each spike. \fref{fig:real}(a) shows the PCA features on the first two principal components, where only 1 cluster can be identified. \fref{fig:real}(b) shows the best 2-D projection of wavelet features used in \texttt{Wave\_clus}, where 2 clusters are identified by the algorithm (average waveforms in \fref{fig:real}(d)). \fref{fig:real}(c) shows the results of clustering using Algorithm~1, where 3 clusters are detected (average waveforms in \fref{fig:real}(e)). Algorithm~2 also detected 3 clusters whose average spike  waveforms are shown in \fref{fig:real}(f) with the corresponding inter-spike interval histograms in \fref{fig:real}(g) which indicate the validity of the extracted clusters.

Regardless of the clustering algorithm used, the methods that rely on PCA for feature extraction usually fail to identify clusters with highly overlapping features in the PCA space, thus assigning spikes generated from different neurons to the same cluster \cite{quiroga_unsupervised_2004,adamos_performance_2008}. We have observed this in several \invivo recordings when the spike waveforms of different neurons are very similar. DWT has been shown more effective for spike feature extraction which provides better separability of clusters. However, similar to PCA, it does not necessarily provide the most discriminative feature subspace for clustering, and may fail to correctly discriminate spikes with similar spike waveform when the noise level is high.

\section{Conclusion} \label{sec:conclusion}
In this paper, we have proposed two unsupervised spike sorting algorithms based on simultaneous discriminative subspace learning and clustering. Both algorithms can detect the number of clusters automatically. The first algorithm keeps increasing the number of clusters and learns the discriminative subspace until no extra separate and significant cluster can be identified. The second algorithm utilizes a divisive clustering scheme which starts by dividing the data samples into two clusters in the most discriminative 1-dimensional projection, and keeps dividing the resultant clusters in two until achieving an almost unimodal distribution in the subspace for each final cluster. This provides better convergence behavior and improved handling of uneven cluster sizes and outliers. We evaluated our algorithms against two commonly used spike sorting methods PCA-$k$means and \texttt{Wave\_clus}, using both synthetic and \invivo data. When tested on synthetic data, the proposed algorithms achieved significantly higher accuracies in all the cases. Furthermore, the results indicate that the algorithms are highly robust to noise. When tested on \invivo data, our algorithms provided better cluster separability compared with PCA and DWT, and can find clusters which may not be identifiable in PCA and wavelet space. Through utilizing adaptive subspace learning we have combined to a single stage the three important parts of conventional spike sorting methods including feature extraction, dimensionality reduction and clustering, which resulted in a significant improvement in spike sorting performance.


A few aspects of our proposed methods require further investigation. First, since the methods are based on \kmeans, they are prone to local optima, and may have problem handling very small sized clusters. Second, the methods require batch processing of spike waveforms and are only suitable for offline setting. It is, however, of practical interest to develop the online versions of the algorithms for real-time spike sorting.

\bibliographystyle{IEEEtran}
\bibliography{refs}

\end{document}